%% file: main.tex
\documentclass[10pt,journal,compsoc]{IEEEtran}
\usepackage{amsmath,amsfonts}
\usepackage{algorithmic}
\usepackage{algorithm}
\usepackage{array}
\usepackage{textcomp}
\usepackage{multirow}
\usepackage{stfloats}
\usepackage{float}
\usepackage{url}
\usepackage{verbatim}
\usepackage{graphicx}
\usepackage{cite}
\usepackage[dvipsnames,table]{xcolor}
\usepackage{etoolbox}
\usepackage{longtable}
\usepackage{pdflscape}
\usepackage{graphicx}
\usepackage{siunitx}
\usepackage{booktabs}
\usepackage{amssymb}
\usepackage{bm}
\usepackage{utfsym}
\usepackage{colortbl}%
\usepackage[pagebackref=true,breaklinks=true,colorlinks,bookmarks=false,citecolor=blue,linkcolor=blue,urlcolor=blue]{hyperref}

\input{preamble}

\begin{document}

\title{Learning Action Manifold with Multi-view Latent Priors for Robotic Manipulation}


\author{Junjin~Xiao$^*$,
        Dongyang~Li$^*$,
        Yandan~Yang,
        Shuang~Zeng,
        Tong~Lin,
        Xinyuan~Chang,
        Feng~Xiong,
        Mu~Xu,
        Xing~Wei,
        Zhiheng~Ma,
        Qing~Zhang,
        Wei-Shi~Zheng
      
\IEEEcompsocitemizethanks{
\IEEEcompsocthanksitem Junjin Xiao, Dongyang Li, Qing Zhang, and Wei-Shi Zheng are with the Key Laboratory of Machine Intelligence and Advanced Computing, Ministry of Education, China. 
\IEEEcompsocthanksitem Yandan Yang, Xinyuan Chang, Feng Xiong and Mu Xu are with AMap, Alibaba Group. 
\IEEEcompsocthanksitem Shuang Zeng, Tong Lin and Xing Wei are with Xi'an Jiaotong University. 
\IEEEcompsocthanksitem Zhiheng Ma is with Shenzhen University of Advanced Technology. 
\IEEEcompsocthanksitem $*$: Equal Contribution.
}
}



\IEEEtitleabstractindextext{%
\input{sec/sec0_abstract}
}

\maketitle

\IEEEdisplaynontitleabstractindextext
\IEEEpeerreviewmaketitle

\input{sec/sec1_intro}

\input{sec/sec2_related_work}

\input{sec/sec3_method}

\input{sec/sec4_experiments}
\input{sec/sec5_conclusion}





\ifCLASSOPTIONcaptionsoff
  \newpage
\fi

\bibliographystyle{IEEEtran}
\bibliography{references}

\end{document}

%% file: preamble.tex
\usepackage[dvipsnames]{xcolor}
\usepackage{subcaption}
\usepackage[subfigure]{tocloft}

\newif\ifdrafting
\draftingtrue 
\ifdrafting
    \newcommand{\todo}[1]{{\leavevmode\color[rgb]{1,0,0}[TODO: #1]}}
    
    \newcommand{\ds}[1]{{\leavevmode\color[rgb]{0,0,1}[DS: #1]}}

    \definecolor{tabfirst}{rgb}{1, 0.7, 0.7} 
    \definecolor{tabsecond}{rgb}{1, 0.85, 0.7} 
    \definecolor{tabthird}{rgb}{1, 1, 0.7} 
\else
    \newcommand{\todo}[1]{}
    \newcommand{\ds}[1]{}
    \newcommand{\mh}[1]{}
    
\fi

\newcommand{\afterfig}{\vspace{-1.25em}}

\newcommand{\aftertabcaption}{\vspace{-0.75em}}

\definecolor{lightblue}{rgb}{0.95,0.98,1.0}
\definecolor{lightgray}{rgb}{0.95,0.95,0.95}

%% file: sec/sec0_abstract.tex
\begin{abstract}
\label{sec:abstract}
This paper addresses the challenges of spatial perception and manipulation in Vision-Language-Action (VLA) models within complex environments. 
While recent works have attempted to enhance VLA perception ability by injecting or distilling features from 3D foundation models, they, however, achieve only limited spatial understanding accuracy improvement, due to biased geometric predictions led by inherent depth ambiguity under monocular input conditions. Furthermore, existing action generation methods rely on indirect paradigms that predict high-dimensional noise or velocity. Regressing these unstructured targets imposes a significant optimization burden, which intensifies as the action dimensionality increases, thereby hindering efficient learning of complex robotic policies. To address these issues, we present a VLA framework with the following two novel designs. First, to tackle the depth ambiguity from monocular input, we propose to leverage pre-trained multi-view diffusion model to synthesize novel views in latent space, obtaining enriched scene context with largely reduced geometric uncertainty. To effectively integrate these multi-view latent priors, we present Geometry-Guided Gated Transformer ($\text{G}^3\text{T}$), which is designed to align multi-view latent features under the guidance of monocular 3D geometric prior and selectively aggregate informative views while suppressing noise from occluded regions based on an adaptive gating mechanism. Second, to overcome the optimization inefficiencies, we introduce Action Manifold Learning (AML). Unlike traditional methods that decode abstract noise or velocity, AML shifts the prediction target directly to actions. This enables the policy to focus on learning the intrinsic structure of the valid action manifold, leading to more efficient and robust execution. Extensive evaluations on LIBERO, LIBERO-Plus, RoboTwin 2.0, and real-world robot experiments demonstrate that our method outperforms state-of-the-art baselines in both success rate and robustness. Our project page is available at \url{https://junjxiao.github.io/Multi-view-VLA.github.io/}.

\end{abstract}


%% file: sec/sec1_intro.tex
\ifCLASSOPTIONcompsoc
\IEEEraisesectionheading{\section{Introduction}\label{sec:introduction}}
\else
\section{Introduction}
\label{sec:introduction}
\fi

\IEEEPARstart{T}{he} emergence of Vision-Language-Action (VLA) models marks a significant paradigm shift in robotic manipulation, empowering autonomous agents to bridge the gap between high-level semantic reasoning and low-level physical execution. By leveraging large-scale pre-trained Vision-Language models~\cite{alayrac2022flamingo, liu2023llava, QwenVL}, VLA frameworks have demonstrated remarkable zero-shot generalization and reasoning capabilities across diverse domains, including complex robotic manipulation~\cite{zitkovich2023rt2, kim24openvla, pi0, xiao2025world} and navigation~\cite{zeng2025janusvln, wei2025streamvln}. To successfully perform complex and long-horizon tasks in the real world, a VLA agent must not only interpret intricate language instructions but also possess precise 3D spatial awareness to efficiently predict action trajectories.

However, current VLA models face two fundamental bottlenecks: limited spatial perception ability under monocular conditions and inefficient action learning mechanisms. Most existing methods~\cite{octo_2023,kim2025fine,Pertsch2025fast} rely primarily on 2D images, suffering from inherently limited spatial awareness due to the irreversible loss of depth information during the projection from 3D physical space to 2D image planes. This limitation frequently leads to execution failures in precision-sensitive tasks. As illustrated in Fig.~\ref{fig:teaser}(a) and (b), recent efforts to equip VLAs with 3D awareness can be categorized into two main paths: integrating explicit 3D inputs (e.g., RGB-D)~\cite{li2026pointvla,Goyal2024rvt2} or leveraging spatial features from pre-trained 3D foundation models~\cite{lin2025evo,li2025spatial,wang2025vggt, depthanything3}. However, the above two paths still have their respective limitations. Specifically, utilizing additional explicit 3D inputs requires costly extra hardware, limiting the overall scalability, while adopting spatial features from 3D foundation models faces the fundamental challenge of monocular depth ambiguity. In typical robotic setups restricted to a single RGB camera, recovering full scene geometry is a highly ill-posed problem. Hence, the spatial features injected into VLA models are often noisy and geometrically inconsistent, adversely affecting the reliability of environmental perception.


In addition to spatial perception, current VLA frameworks face inherent limitations in action generation. Prevailing generative approaches~\cite{pi0, Chi2023DiffusionPolicy, gr00tn1, li2024cogact}, such as diffusion~\cite{ho2020denoising,song2020denoising} and flow matching~\cite{lipman2022flow,liu2022flow,geng2025mean}, rely on indirect targets like noise ($\epsilon$-prediction) or velocity ($v$-prediction). However, these targets represent high-dimensional, unstructured statistical fields that lack direct physical semantics. Forcing VLA networks to regress such abstract signals imposes a significant optimization burden, as the model must disentangle complex noise patterns rather than learn meaningful action structures. This learning difficulty becomes more pronounced as the action dimensionality increases (e.g., in multi-arm or whole-body control), where the expanded search space makes it increasingly challenging for indirect prediction methods to converge to robust and precise policies.


\input{Figures/teaser}

In this paper, we present a VLA framework that addresses the above issues encountered by existing VLA models through a synergistic design for enhancing both spatial perception and action learning efficiency. To overcome the depth ambiguity from monocular inputs, we propose to leverage pre-trained multi-view diffusion models to synthesize latent representations of novel views, providing complementary geometric cues from multiple angles and thereby resolving monocular geometric uncertainties. To robustly integrate these multi-view cues, we introduce Geometry-Guided Gated Transformer ($\text{G}^3\text{T}$), which utilizes monocular 3D geometric prior to guide the alignment of multi-view latent features, and embed with an adaptive gating mechanism to selectively aggregate informative views while suppressing noise from occluded regions, ensuring that the VLA model receives reliable and geometrically consistent spatial embeddings. To avoid the limitation from indirect action generation, we propose Action Manifold Learning (AML) based on the ``Action Manifold Hypothesis'': successful actions are not randomly scattered but reside on a low-dimensional, smooth manifold shaped by physics, task goals, and environmental constraints (as shown in Fig.~\ref{fig:aml}). Unlike traditional diffusion policies that predict noise or velocity, AML directly predicts clean action chunks on this underlying action manifold. By explicitly mapping policy outputs to action trajectories, our approach eliminates the inefficiency of indirect decoding, ensuring more efficient optimization.

We evaluate our method on several benchmarks, including LIBERO~\cite{liu2023libero}, LIBERO-Plus~\cite{fei2025libero_plus}, RoboTwin 2.0~\cite{chen2025robotwin}, and real-world robotic tasks. Experimental results demonstrate that our framework consistently outperforms state-of-the-art methods in both success rate and robustness.

Our main contributions are summarized as follows:
\begin{itemize}
    \item We present a VLA framework that enables reliable spatial perception and efficient action learning, allowing for robust and precise robotic manipulation. 
   
    \item We introduce Geometry-Guided Gated Transformer to address the inherent monocular depth ambiguity by leveraging multi-view diffusion priors to provide geometry guidance. 
    
    \item We propose Action Manifold Learning, a direct action prediction mechanism to avoid the limitations of traditional diffusion-based indirect noise/velocity decoding, achieving more efficient action learning.
\end{itemize}

%% file: Figures/teaser.tex
\begin{figure*}[t]
\begin{center}
\includegraphics[width=1\textwidth]{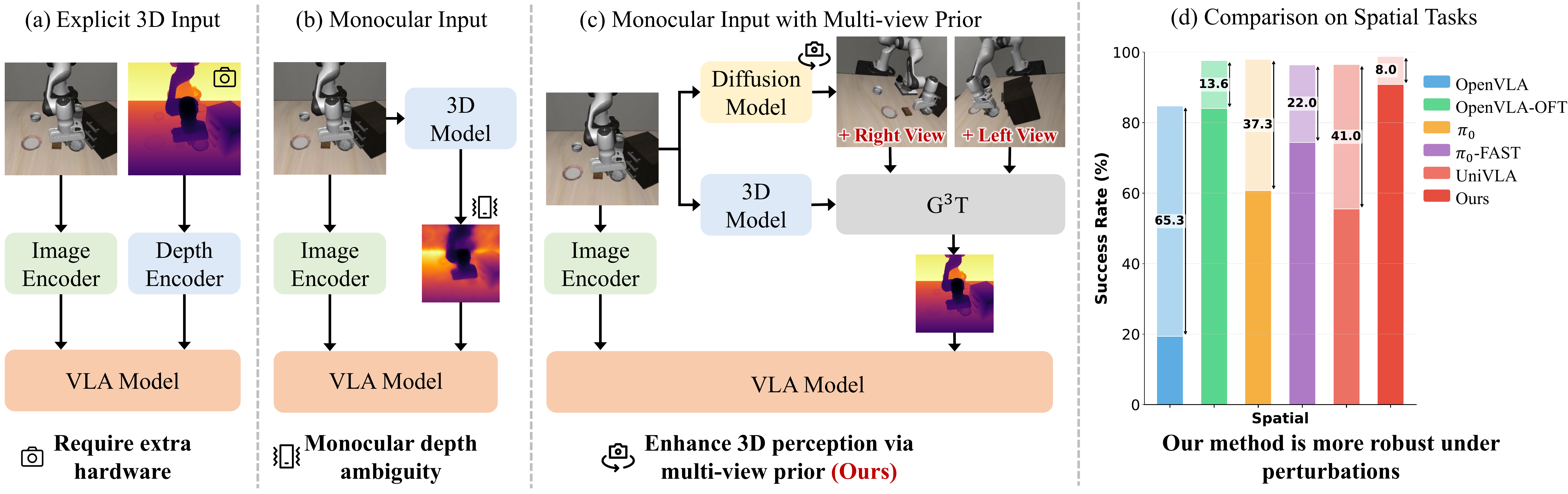} \\
\end{center}
\afterfig{}
\caption{\textbf{Methodology comparison.} Existing VLA models either rely on expensive RGB-D sensors for explicit 3D input (a) or suffer from severe depth ambiguity under monocular setting (b). In contrast, our method leverages multi-view diffusion prior and Geometry-Guided Gated Transformer ($\text{G}^3\text{T}$) to synthesize robust geometric features from a single RGB image, resolving the depth ambiguity without utilizing extra hardware (c). As shown in (d), our method demonstrates stability against disturbances under LIBERO-Spatial tasks. Dark bars: success rate under perturbation, light bars: original result. Our approach exhibits minimal degradation (8.0) compared to baselines.}
\label{fig:teaser}
\end{figure*}

%% file: sec/sec2_related_work.tex
\section{Related Work}
\label{sec:related}

\subsection{Vision-Language-Action Models}
The prevailing paradigm in Vision-Language-Action (VLA) research leverages pre-trained Vision-Language Models (VLMs)~\cite{alayrac2022flamingo,li2022blip,li2023blip2,dai2023instructblip,liu2023llava,liu2023improvedllava,QwenVL,Karamcheti2024prismatic} as backbone, adapting them to robotic control via large-scale demonstration data~\cite{zitkovich2023rt2,octo_2023,kim24openvla,kim2025fine,zheng2025xvla,ye2026st4vla,zhao2025cotvla,bu2025univla,cen2025worldvla,song2025reconvla}. Early approaches primarily rely on Supervised Fine-Tuning (SFT). RT-2~\cite{zitkovich2023rt2} pioneers the formulation of action generation as autoregressive token prediction, while OpenVLA~\cite{kim24openvla} establishes a strong open-source baseline by fine-tuning Prismatic VLM~\cite{Karamcheti2024prismatic}. To mitigate the latency of sequential decoding, recent works such as $\pi_0$~\cite{pi0} and Octo~\cite{octo_2023} employ diffusion or flow-matching heads for parallel continuous action generation, offering superior temporal resolution. However, SFT-based methods are inherently limited by distributional shift and error accumulation when faced with out-of-distribution scenarios.

To address these limitations, Reinforcement Learning (RL) is increasingly integrated into VLA frameworks for post-training refinement. Model-free approaches, such as RIPT~\cite{tan2025ript}, utilize lightweight, critic-free advantage estimation to efficiently refine policies in few-shot regimes without compromising pre-trained priors~\cite{Kalashnikov2018scalable,Kumar2023ptr,bhateja2023robot,nakamoto2024steering,xu2025rldg,Johannink2019Residual,mark2024policy,lu2025vlarl,chen2025conrft,li2025simplevla}. Alternatively, model-based methods like World-Env~\cite{xiao2025world} construct latent world model to enable risk-free, data-efficient policy optimization, extending the Dreamer series' paradigm~\cite{hafner2019dreamer,hafner2020mastering,hafner2025dreamerv3} to embodied AI~\cite{hansen2023td,WMPO2025,fei2025srpo,li2025vla,jiang2026wovr}. These RL-enhanced strategies significantly improve robustness and generalization beyond the capabilities of pure imitation learning.

\input{Figures/aml}

\subsection{3D Spatial Perception in VLA}
While 2D VLMs excel at semantic reasoning, they basically lack explicit geometric grounding, limiting their precision in manipulation tasks. To address this problem, recent efforts propose to enhance 3D perception by injecting geometric features from point clouds~\cite{li2026pointvla,Ze2024DP3,wang2024vihe}, depth maps~\cite{bhat20253d,yuan2025depthvla,sun2025geovla,shi2025spatialactor}, or voxels~\cite{liu2025vox}. For instance, DP3~\cite{Ze2024DP3} processes reconstructed point clouds via PointNet~\cite{Charles2017pointnet}, while DepthVLA~\cite{yuan2025depthvla} fuses depth information via cross-attention. Other works leverage 3D foundation models like VGGT~\cite{wang2025vggt} to extract rich geometric priors, fused through feature projection~\cite{abouzeid2025geoaware} or gated mechanisms~\cite{lin2025evo,yu20263dmixvl}. However, these methods rely on auxiliary depth sensors or complex reconstruction pipelines, increasing hardware costs. Furthermore, single-view inputs often suffer from inherent depth ambiguity and occlusion, leading to unreliable geometric priors.

To mitigate these issues, some approaches introduce explicit spatial alignment or active view selection, such as RVT-2~\cite{Goyal2024rvt2} and BridgeVLA~\cite{li2025bridgevla}, which utilize multi-view data or dynamic camera control~\cite{Bai_2026_CVPR}. In contrast, our method achieves robust 3D spatial perception using only a single RGB image. By synthesizing geometrically consistent multi-view cues and employing an adaptive gating mechanism, we resolve depth ambiguity without requiring additional hardware or multi-view coordination, offering a more practical and scalable solution for real-world deployment.

\subsection{Generative Action Prediction in VLA}
Generative modeling, particularly Denoising Diffusion Probabilistic Models (DDPM)~\cite{ho2020denoising}, DDIM~\cite{song2020denoising}, and Flow Matching (FM)~\cite{lipman2022flow,liu2022flow}, revolutionizes action prediction in VLA architectures. These paradigms offer superior modeling of complex, multimodal action distributions compared to conventional MSE regression or autoregressive decoding, naturally supporting action chunking for precise temporal coordination~\cite{Chi2023DiffusionPolicy,octo_2023,Ze2024DP3,pi0,li2024cogact,gr00tn1}. 

Diffusion Policy~\cite{Chi2023DiffusionPolicy} pioneers conditional diffusion for visuomotor control, demonstrating significant improvements over deterministic baselines. More recently, $\pi_0$~\cite{pi0} employs a flow-matching action expert based on a DiT architecture for efficient trajectory generation via ODE integration. Similarly, GR00T N1~\cite{gr00tn1} couples a deliberative VLM backbone with a fast flow-matching generator, while CogACT~\cite{li2024cogact} aligns diffusion-based control with cognitive reasoning. Despite their success, these generative methods often struggle with high-dimensional action spaces due to error accumulation during iterative sampling. Our proposed Action Manifold Learning (AML) module addresses this by structuring the optimization landscape, enabling accurate action prediction even with limited sampling steps.

%% file: Figures/aml.tex
\begin{figure}[t]
\begin{center}
\includegraphics[width=1\linewidth]{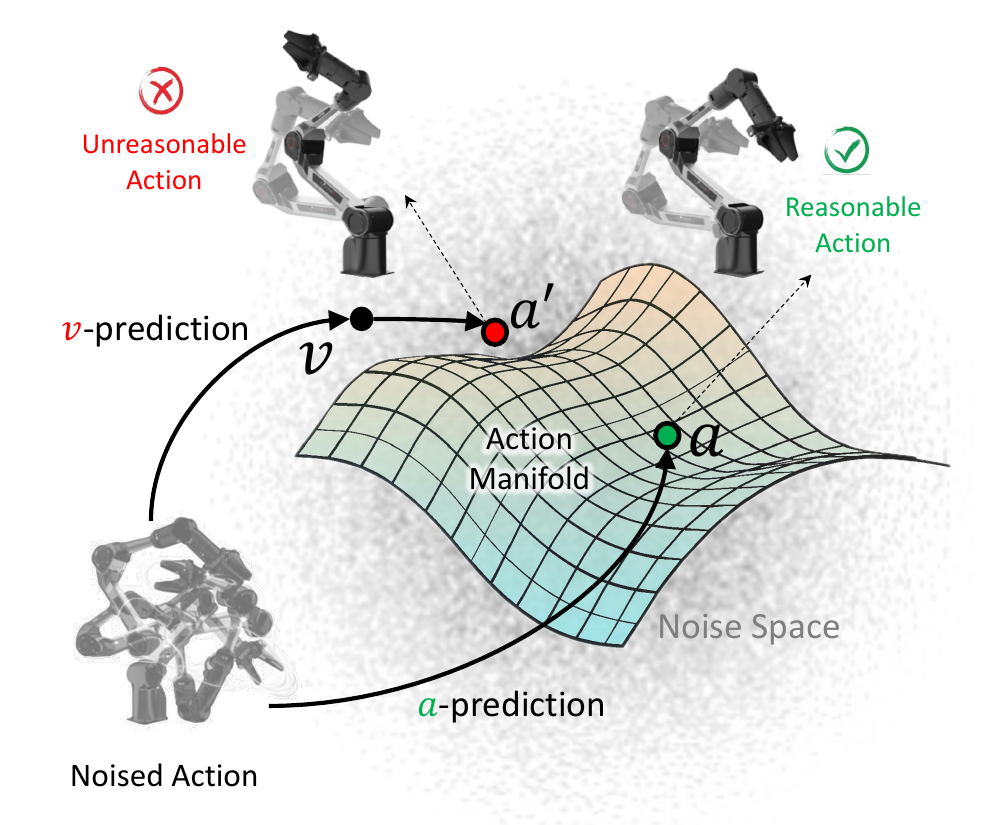} \\
\end{center}
\afterfig{}
\caption{\textbf{Action manifold hypothesis.} We posit that a meaningful action sequence is a highly structured entity residing
on a low-dimensional action manifold. The conventional prediction targets of noise or velocity are inherently high-dimensional and off-manifold, which increase the burden of model learning and lead to unreasonable action.}
\label{fig:aml}
\end{figure}

%% file: sec/sec3_method.tex
\section{Method}
\label{sec:method}
\input{Figures/method}

We first outline the overall framework in Sec.~\ref{subsec:overall}, detailing the end-to-end pipeline from multimodal inputs to action outputs. Next, Sec.~\ref{subsec:g^3t} introduces the Geometry-Guided Gated Transformer ($\text{G}^3\text{T}$), which aligns multi-view latent features into spatially consistent embeddings and adaptively filters out occluded views. Then, Sec.~\ref{subsec:aml} presents Action Manifold Learning, a direct prediction mechanism that enhances optimization efficiency. Finally, we describe our training and implementation details in Sec~\ref{subsec:implementation}.

\subsection{Overall Framework}
\label{subsec:overall}
The overview of our method is given in Fig.~\ref{fig:method}. Our framework adopts a modular architecture that decouples high-level semantic perception from low-level action execution, consisting of two primary components: a Vision-Language Model (VLM) serving as the perception engine, and an Action Expert based on our proposed Action Manifold Learning (AML) for decision-making. This design leverages the robust reasoning capabilities of large-scale VLMs while ensuring precise and stable control through specialized geometric-aware action modeling. The framework processes a multimodal input stream consisting of a primary third-person view image $I_{main}$, a natural language instruction $L$, and the current robot state $q$. An optional wrist-mounted camera view $I_{wrist}$ can be incorporated to provide egocentric details. The final output is a sequence of continuous action commands $\hat{A}_t = [\hat{a}_t, \dots, \hat{a}_{t+H-1}]$ over a horizon $H$, which directly controls the robot actuators.

\vspace{0.5em}
\noindent \textbf{Semantic feature extraction.} 
For semantic understanding, we employ Qwen3-VL~\cite{bai2025qwen3vltechnicalreport} as the backbone VLM due to its superior image-text alignment and long-context modeling capabilities. The main view $I_{main}$ (along with the optional $I_{wrist}$) and the instruction $L$ are tokenized and processed by the VLM encoder $\mathcal{E}_{vlm}$ to extract high-level semantic feature $\phi_{sem}$ from the last layer of VLM:
\begin{equation}
    \phi_{sem} = \mathcal{E}_{vlm}(I_{main}, I_{wrist}, L).
\end{equation}
These features capture the semantic intent and object relationships but lack precise metric spatial information.


\vspace{0.5em}
\noindent \textbf{Geometry-enhanced spatial perception.} 
To complement semantic representations with robust spatial awareness, we introduce a geometry-enhanced perception module that utilizes explicit geometric priors with synthesized multi-view cues. To this end, we first extract dense 3D structural prior from the main view $I_{main}$, utilizing the spatial encoder of VGGT~\cite{wang2025vggt} to process $I_{main}$ and extracting the hidden state from its last layer:
\begin{equation}
    \phi_{vggt} = \mathcal{E}_{vggt}(I_{main}).
\end{equation}
As VGGT is trained on large-scale 3D datasets, these high-level features implicitly encode rich geometric information, providing strong global spatial context even from a single monocular input. However, relying solely on monocular feature can be susceptible to severe occlusions and depth ambiguities in complex manipulation scenarios. To alleviate the depth ambiguities from monocular input, we propose to synthesize complementary multi-view information using the 6B LongCat-Image-Edit model~\cite{LongCatImage}. This multi-view synthesis operates entirely in the latent space rather than the pixel space, which is elaborated in the following. 

Let $\mathcal{E}_{vae}$ denote the encoder of a pre-trained Variational Autoencoder (VAE). We first encode the main view into a latent representation $z_{main} = \mathcal{E}_{vae}(I_{main})$. Then, a Diffusion Transformer (DiT) denoiser $\mathcal{D}_{dit}$, conditioned on $z_{main}$ and target view prompts ($p_{l}, p_{r}$), is employed to generate the latent representations of two novel virtual views $(z_{l}, z_{r})$:
\begin{equation}
    z_{novel} = \mathcal{D}_{dit}(z_{main}, p_{target}), \quad \text{where } z_{novel} \in \{z_{l}, z_{r}\}.
\end{equation}
Operating in latent space rather than pixel space offers two key advantages. First, it drastically improves computational efficiency by allowing us to reduce the latent resolution to $256 \times 256$ and limit the denoising steps to 2, as we bypass the computationally expensive VAE decoder. Second, and more importantly, it enhances the overall robustness. By performing synthesis at the feature level, we avoid introducing pixel-level artifacts or physically implausible details (e.g., distorted textures or inconsistent lighting) that often plague image-generation models. This ensures that the downstream VLA model receives clean, semantically consistent geometric cues rather than noisy visual distractions. Since this generative module is frozen during policy training, we pre-compute and cache the latent features $\{z_{l}, z_{r}\}$ for the entire dataset, further accelerating the training pipeline.

Finally, the latent features of the synthesized multi-views and the geometric embeddings from VGGT are fused by our proposed Geometry-Guided Gated Transformer ($\text{G}^3\text{T}$) to align these heterogeneous features into a unified, geometrically consistent spatial embedding $\phi_{geo}$, while adaptively gating out noisy or occluded information by:
\begin{equation}
    \phi_{geo} = \text{G}^3\text{T}(z_{l}, z_{r}, \phi_{vggt}).
\end{equation}

\vspace{0.5em}
\noindent \textbf{Feature fusion and action prediction.} 
To integrate semantic information with geometric perception, we fuse the semantic features $\phi_{sem}$ and the enhanced spatial embeddings $\phi_{geo}$ via a standard cross-attention mechanism. Specifically, we project $\phi_{sem}$ and $\phi_{geo}$ into query, key, and value spaces using learnable linear transformations $W_Q$, $W_K$, and $W_V$, respectively. The attention scores are computed by measuring the compatibility between the semantic queries and spatial keys, which are then used to weight the spatial values. Finally, we use an output projection matrix $W_O$ to map the aggregated features back to the model dimension. The fusion process is formulated as:
\begin{equation}
    \phi = W_O \cdot \text{Softmax}\left(\frac{Q K^T}{\sqrt{d_k}}\right) V,
\end{equation}
where $Q = W_Q \phi_{sem} $, $K = W_K\phi_{geo} $, and $V = W_V\phi_{geo} $. Here, $d_k$ denotes the dimension of the key vectors, and $W_Q, W_K, W_V, W_O$ are learnable parameters. This design allows the semantic context to dynamically attend to the most relevant geometric structures, resulting in a comprehensive multimodal representation.

The resulting comprehensive context representation $\phi$, combined with the current robot state, is passed to the Action Expert. Unlike traditional diffusion policies that indirectly predict noise or velocity, our Action Expert employs the Action Manifold Learning (AML) mechanism (Sec.~\ref{subsec:aml}) to directly map the multimodal context onto a low-dimensional action manifold, predicting the action chunk $\hat{A}_t$. This ensures that the generated actions are semantically aligned and geometrically grounded.

\input{Figures/g3t}

\subsection{Geometry-Guided Gated Transformer}
\label{subsec:g^3t}

The Geometry-Guided Gated Transformer ($\text{G}^3\text{T}$) is designed to fuse heterogeneous spatial features, specifically the monocular geometric prior from VGGT and the synthesized multi-view latent representations, into a unified and robust spatial embedding $\phi_{geo}$. The detailed illustration is shown in Fig.~\ref{fig:g3t}. The module operates through three sequential stages: cross-view alignment, adaptive gated fusion, and geometric consistency refinement.

\vspace{0.5em}
\noindent \textbf{Cross-view alignment.} 
Let $\phi_{vggt} \in \mathbb{R}^{N \times C_v}$ denote the spatial feature extracted from the main view by VGGT, where $N$ is the number of spatial tokens and $C_v$ is the feature dimension. Let $z_{l}, z_{r} \in \mathbb{R}^{M \times C_z}$ be the latent features of the synthesized left and right views, respectively, where $M$ is the number of latent tokens and $C_z$ is the latent dimension. Typically, $C_v \neq C_z$ and $N \neq M$. To align these heterogeneous features into a shared latent space, we employ learnable linear projections. Specifically, we use a dedicated projector $W_{v} \in \mathbb{R}^{C_v \times C}$ for the VGGT feature, and a shared projector $W_{z} \in \mathbb{R}^{C_z \times C}$ for both multi-view features:
\begin{equation}
    \tilde{\phi}_{vggt} = W_{v}\phi_{vggt} , \quad \tilde{z}_{l} =  W_{z}z_{l}, \quad \tilde{z}_{r} = W_{z}z_{r} .
\end{equation}
Here, $C$ denotes the unified hidden dimension. We then concatenate these projected features along the token dimension to form a joint representation $\phi_{concat} = [\tilde{\phi}_{vggt}; \tilde{z}_{l}; \tilde{z}_{r}] \in \mathbb{R}^{(N+2M) \times C}$. 

Subsequently, we apply a Multi-Head Self-Attention (MHSA)~\cite{Vaswani2017transformer} layer to $\phi_{concat}$:
\begin{equation}
    \phi_{aligned} = \text{MHSA}(\phi_{concat}) + \phi_{concat}.
\end{equation}
This cross-view alignment mechanism serves two pivotal roles. First, it enables interaction between the monocular VGGT feature and the synthesized multi-view features. By leveraging the structural information encoded in the novel view representations $\tilde{z}_{l}$ and $\tilde{z}_{r}$, the model effectively resolves the depth ambiguity inherent in the single-view $\tilde{\phi}_{vggt}$. Second, it facilitates attention between the left and right views themselves. This inter-view interaction acts as a mutual regularization process, where consistent structural cues are reinforced while high-frequency noise artifacts, resulting from the limited denoising steps in synthesis, are suppressed. After this step, we split $\phi_{aligned}$ back into aligned components denoted as $\phi'_{vggt} \in \mathbb{R}^{N \times C}$, $z'_{l} \in \mathbb{R}^{M \times C}$, and $z'_{r} \in \mathbb{R}^{M \times C}$.

\vspace{0.5em}
\noindent \textbf{Adaptive gated fusion.} 
Although alignment improves feature quality, direct utilization of synthesized views may still propagate errors from severely occluded regions. To address this, we introduce an adaptive gating mechanism that dynamically weighs the contribution of each synthesized view based on its reliability. Specifically, we concatenate the aligned left and right features $[z'_{l}; z'_{r}]$ and pass them through a lightweight Multi-Layer Perceptron (MLP) followed by a Sigmoid activation function to predict a scalar gate value for each token. This results in a gate map $G \in \mathbb{R}^{M \times 1}$:
\begin{equation}
    G = \sigma(\text{MLP}([z'_{l}; z'_{r}])),
\end{equation}
where $\sigma(\cdot)$ denotes the sigmoid function. The gate $G$ represents the confidence score for the left view feature at each spatial location. Consequently, the confidence for the right view is implicitly represented as $(1-G)$. We then compute the fused multi-view feature $z_{fused} \in \mathbb{R}^{M \times C}$ via element-wise weighted summation, where $G$ is broadcasted across the feature dimension:
\begin{equation}
    z_{fused} = G \odot z'_{l} + (1 - G) \odot z'_{r},
\end{equation}
where $\odot$ denotes the element-wise multiplication with broadcasting. This mechanism allows the model to selectively aggregate informative regions from the left view when the right view is occluded, or vice versa, effectively filtering out invalid geometric cues caused by occlusions.

\vspace{0.5em}
\noindent \textbf{Geometric Consistency Refinement.} 
Finally, to integrate the fused multi-view information with the original monocular geometry, we concatenate the refined VGGT feature $\phi'_{vggt}$ and the fused multi-view features $z_{fused}$ along the token dimension to form a combined sequence $[\phi'_{vggt}; z_{fused}] \in \mathbb{R}^{(N+M) \times C}$. We apply a final MHSA layer to allow global interaction among all spatial tokens, producing the final geometry-enhanced spatial embedding:
\begin{equation}
    \phi_{geo} = \text{MHSA}([\phi'_{vggt}; z_{fused}]) + [\phi'_{vggt}; z_{fused}].
\end{equation}
The output $\phi_{geo} \in \mathbb{R}^{(N+M) \times C}$ captures both the explicit 3D prior from VGGT and the implicit multi-view consistency, serving as the robust spatial context for subsequent action prediction.

\input{Tables/libero_plus}
\input{Tables/libero}
\input{Tables/robotwin}

\subsection{Action Manifold Learning}
\label{subsec:aml}

Conventional diffusion or flow-based generative models are typically trained to predict noise ($\epsilon$-prediction) or velocity ($v$-prediction). We argue that this indirect formulation poses a fundamental limitation for robotic learning. According to the manifold hypothesis~\cite{SSL,carlsson2009topology}, high-dimensional real-world data, such as natural images and human language, do not scatter randomly but rather reside on intrinsic low-dimensional manifolds. We extend this insight to robotics, positing that coherent and meaningful robot action sequences also constitute highly structured entities lying on a low-dimensional \textit{action manifold}. In contrast, the prediction targets of noise or velocity are inherently high-dimensional and often lie off this manifold~\cite{li2025jit,vincent2010stacked}. Forcing a network with finite capacity to regress these unstructured, high-dimensional targets is inefficient, as it expends significant model capacity on filtering ambient noise rather than learning the underlying action semantics. This challenge escalates significantly as embodiment complexity increases (e.g., from single-arm manipulators to full-body humanoids), leading to exponentially larger action spaces. To alleviate this optimization burden, we propose Action Manifold Learning, which shifts the prediction target from noise/velocity to the action itself (denoted as $a$-prediction). This allows the action expert to focus directly on learning the intrinsic structure and semantics of valid actions.

We employ the Diffusion Transformer~\cite{Peebles2022DiT} (DiT) as our action generator, denoted as $V_\theta$. Instead of predicting the score function or velocity field directly, $V_\theta$ predicts the denoised action chunk $\hat{A}_t = [\hat{a}_t, \hat{a}_{t+1}, \dots, \hat{a}_{t+H-1}]$ of horizon $H$. Given a ground-truth action chunk $A_t$, a diffusion timestep $\tau \in [0, 1]$, and noise $\epsilon \sim \mathcal{N}(0, \mathbf{I})$, we construct the noisy action input as $A^\tau_t = \tau A_t + (1 - \tau) \epsilon$. The network $V_\theta$ takes the multimodal feature $\phi_t$, the current robot state $q_t$, and the noisy action $A^\tau_t$ as inputs, and directly outputs the estimated clean action chunk $\hat{A}_t$:
\begin{equation}
    \hat{A}_t = V_\theta(\phi_t, A^\tau_t, q_t).
    \label{eq:action_pred}
\end{equation}

Although the model explicitly predicts the clean action $\hat{A}_t$, we optimize the network using a velocity-consistent loss, which empirically yields superior stability and convergence compared to direct action regression. Specifically, we derive the estimated velocity $\hat{v}$ and the ground-truth velocity $v$ based on the linear interpolation trajectory:
\begin{equation}
    \hat{v} = \frac{\hat{A}_t - A^{\tau}_t}{1-\tau},
    v = \frac{A_t - A^{\tau}_t}{1-\tau}.
\label{eq:velocity_def}
\end{equation}
We then minimize the Mean Squared Error (MSE) between the predicted and ground-truth velocities. This is equivalent to minimizing a reweighted action loss:
\begin{equation}
    \mathcal{L}(\theta) = \mathbb{E}_{\tau, \epsilon} \left[ w(\tau) \| V_\theta(\phi_t, A^\tau_t, q_t) - A_t \|^2 \right],
    \label{eq:loss_final}
\end{equation}
where the weighting function is $w(\tau) = \frac{1}{(1 - \tau)^2}$. This weight arises from the Jacobian of the transformation from the action space to the velocity space. It elegantly preserves the advantages of flow matching by dynamically adjusting the learning signal strength across noise levels. Intuitively, as $\tau \to 1$ (low noise), the weight increases, compelling the model to perform fine-grained refinements. Conversely, for small $\tau$ (high noise), the lower weight allows the model to focus on coarse structural corrections.

\vspace{0.5em}
\noindent \textbf{Inference process.} During inference, we generate actions by solving the Ordinary Differential Equation (ODE) defined by the learned velocity field. Starting from pure noise $A^0_t \sim \mathcal{N}(0, \mathbf{I})$, we perform iterative denoising over $N$ steps. At each timestep $\tau$, the model predicts the clean action $\hat{A}_t = V_\theta(\phi_t, A^\tau_t, q_t)$ via Eq.~\eqref{eq:action_pred}. We then compute the instantaneous flow velocity $\hat{v}$ using Eq.~\eqref{eq:velocity_def}. Finally, we update the action state using a numerical integrator (e.g., Euler method):
\begin{equation}
    A^{\tau+\Delta\tau}_t = A^\tau_t + \Delta\tau \cdot \hat{v}.
    \label{eq:ode_update}
\end{equation}
This process combines the semantic clarity of direct action prediction at the model level with the smooth, stable trajectory generation capabilities inherent to flow-based dynamics.

\input{Tables/vlm_feature}
\input{Tables/vlm_fusion}

\input{Tables/g3t}

\input{Tables/ablation}
\input{Tables/depth}
\input{Tables/robust}

\subsection{Implementation Details}
\label{subsec:implementation}


Our model is built on the StarVLA codebase~\cite{community2026starvla} and fine-tuned from~\cite{yang2026abotm0}. We use the AdamW~\cite{2018Decoupled} optimizer with weight decay of $1.0 \times 10^{-8}$. The VLM backbone (Qwen3-VL 4B) uses a learning rate of $1.0 \times 10^{-5}$, while the 16-layer DiT action expert uses $1.0 \times 10^{-4}$ with a cosine scheduler (5k steps warmup). We train the model on 4 NVIDIA H20 GPUs (batch size 16 per GPU, bfloat16) for $30K$ steps, which takes approximately 27 hours. Input images are resized to $224 \times 224$. During inference, we use 4-step action denoising to balance efficiency and quality.

%% file: Figures/method.tex
\begin{figure*}[t]
\begin{center}
\includegraphics[width=1\textwidth]{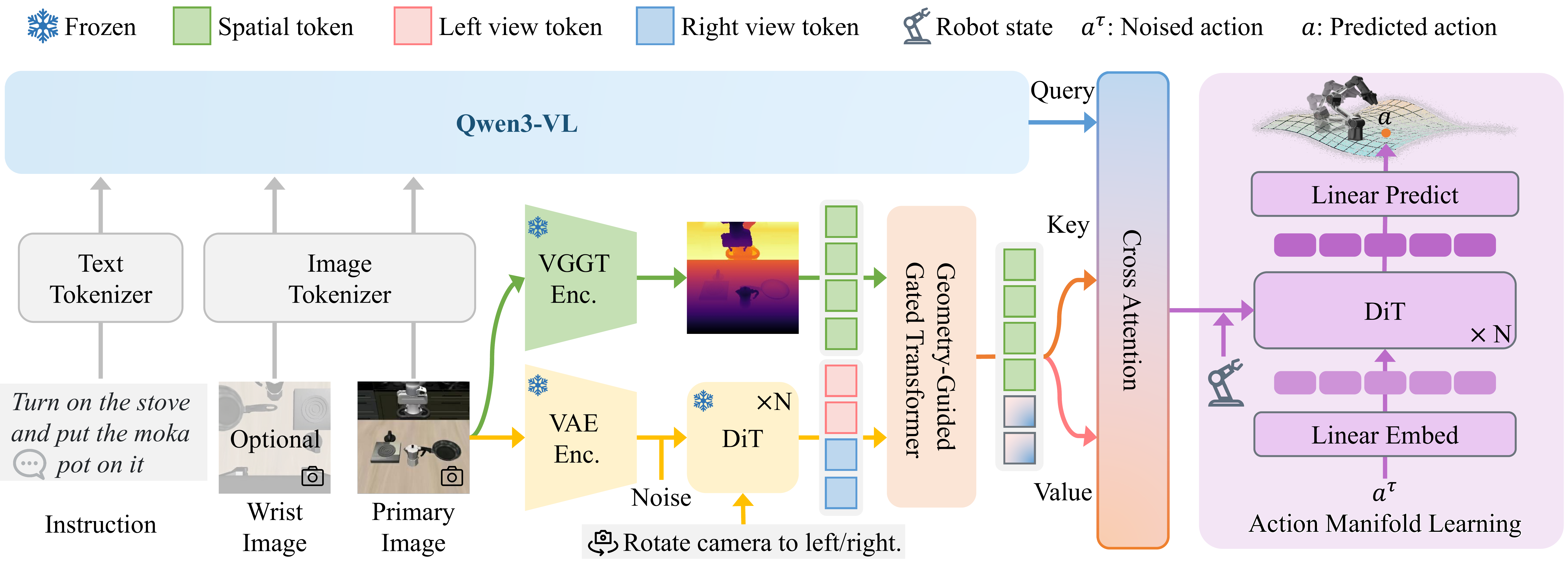} \\
\end{center}
\afterfig{}
\caption{\textbf{Overview of our method.} Our method processes multimodal inputs via a VLM (Qwen3-VL) for semantic features. To enhance spatial awareness, we introduce a geometry module that combines monocular prior from VGGT with multi-view latents synthesized by a diffusion model. These are fused by our Geometry-Guided Gated Transformer ($\text{G}^3\text{T}$), which aligns features and adaptively gates occlusions to produce robust embeddings. Finally, semantic and geometric features are integrated via cross-attention and passed to the Action Manifold Learning (AML) expert. AML directly predicts action chunks on a low-dimensional manifold using a DiT, ensuring stable and precise control.}
\label{fig:method}
\end{figure*}

%% file: Figures/g3t.tex
\begin{figure}[t]
\begin{center}
\includegraphics[width=1\linewidth]{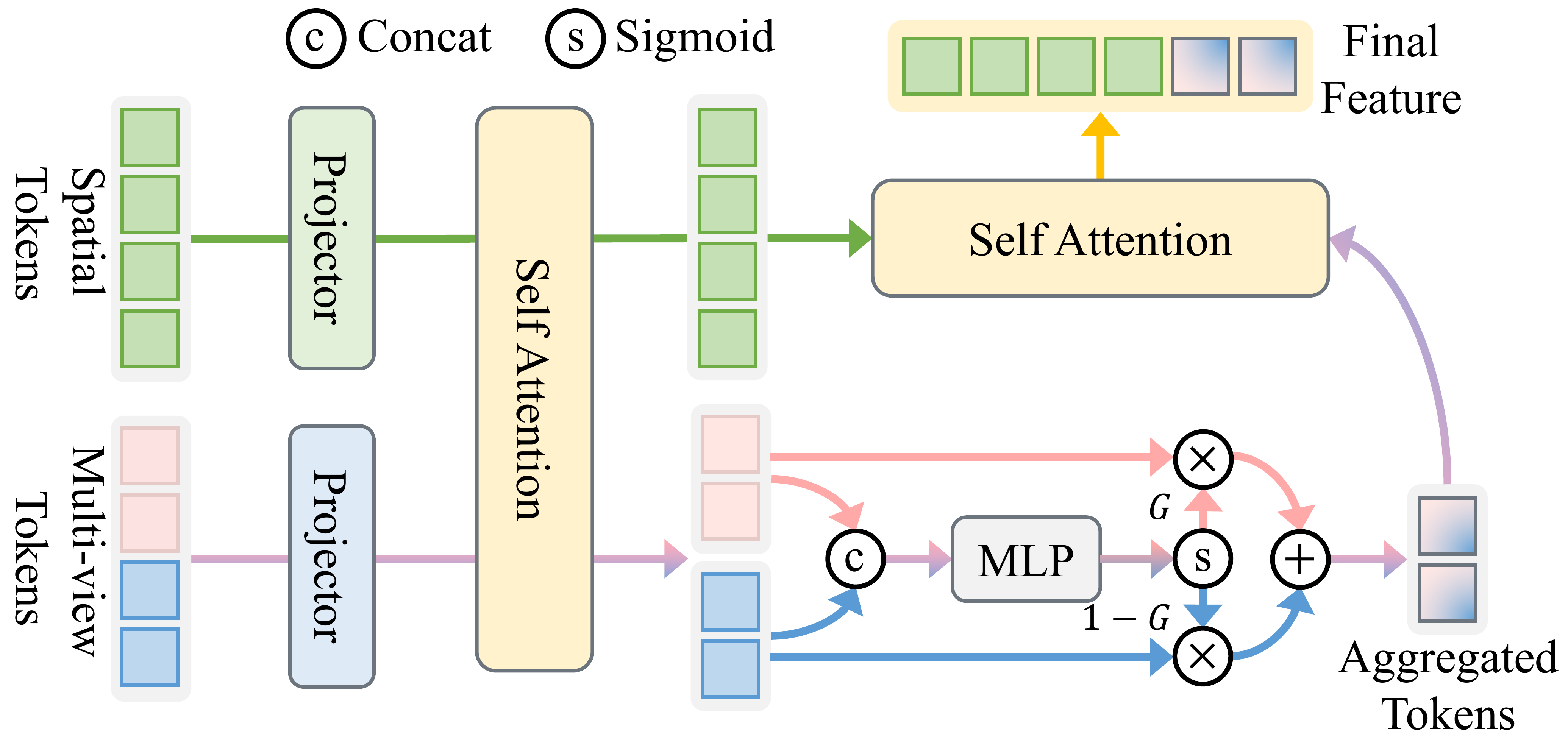} \\
\end{center}
\afterfig{}
\caption{\textbf{Architecture of Geometry-Guided Gated Transformer.} We fuse monocular spatial tokens and synthesized multi-view tokens via $\text{G}^3\text{T}$, producing a robust, occlusion-aware spatial representation.}
\label{fig:g3t}
\end{figure}

%% file: Tables/libero_plus.tex
\begin{table*}[th]
\centering
\caption{\textbf{Zero-shot performance on LIBERO-Plus.} All methods are trained only on the standard LIBERO dataset without fine-tuning on LIBERO-Plus dataset. OpenVLA-OFT\_w shows the performance of OpenVLA-OFT without wrist observation input. OpenVLA-OFT\_m shows the performance of OpenVLA-OFT with a mix-sft~\cite{fei2025libero_plus}.}
\aftertabcaption{}
\label{tab:libero_plus}
\resizebox{0.9\linewidth}{!}{%
\begin{tabular}{lcccccccc}\toprule
\textbf{Method} & \textbf{Camera} & \textbf{Robot} & \textbf{Language} & \textbf{Light} & \textbf{Background} & \textbf{Noise} & \textbf{Layout} & \textbf{Total} \\
\midrule
OpenVLA~\cite{kim24openvla} & 0.8 & 3.5 & 23.0 & 8.1 & 34.8 & 15.2 & 28.5 & 15.6 \\

\rowcolor{lightgray} 
OpenVLA-OFT~\cite{kim2025fine} & 56.4 & 31.9 & 79.5 & 88.7 & 93.3 & 75.8 & 74.2 & 69.6 \\

OpenVLA-OFT\_w~\cite{kim2025fine} & 10.4 & 38.7 & 70.5 & 76.8 & 93.6 & 49.9 & 69.9 & 55.8 \\

\rowcolor{lightgray}
OpenVLA-OFT\_m~\cite{kim2025fine} & 55.6 & 21.7 & 81.0 & 92.7 & 91.0 & 78.6 & 68.7 & 67.9 \\

NORA~\cite{hung2025nora} & 2.2 & 37.0 & 65.1 & 45.7 & 58.6 & 12.8 & 62.1 & 39.0 \\

\rowcolor{lightgray}
WorldVLA~\cite{cen2025worldvla} & 0.1 & 27.9 & 41.6 & 43.7 & 17.1 & 10.9 & 38.0 & 25.0 \\

UniVLA~\cite{bu2025univla} & 1.8 & 46.2 & 69.6 & 69.0 & 81.0 & 21.2 & 31.9 & 42.9 \\

\rowcolor{lightgray}
$\pi_0$~\cite{pi0} & 13.8 & 6.0 & 58.8 & 85.0 & 81.4 & 79.0 & 68.9 & 53.6 \\

$\pi_0$-Fast~\cite{Pertsch2025fast} & 65.1 & 21.6 & 61.0 & 73.2 & 73.2 & 74.4 & 68.8 & 61.6 \\

\rowcolor{lightgray}
RIPT-VLA~\cite{tan2025ript} & 55.2 & 31.2 & 77.6 & 88.4 & 91.6 & 73.5 & 74.2 & 68.4 \\

MergeVLA~\cite{fu2026mergevla} & 50.7 & 30.3 & 66.0 & 84.2 & 85.7 & 66.0 & 68.1 & 62.5 \\

\rowcolor{lightgray}
UnifoLM-VLA-0~\cite{unifolm_vla} & 56.7 & \textbf{69.5} & \textbf{91.2} & 93.6 & 95.3 & 77.9 & 77.9 & 78.9 \\

\midrule
\textbf{Ours} & \textbf{89.6} & 60.1 & 86.9 & \textbf{98.0} & \textbf{95.7} & \textbf{97.2} & \textbf{78.2} & \textbf{85.7} \\
\bottomrule
\end{tabular}
}
\end{table*}

%% file: Tables/libero.tex
\begin{table}[t]
\centering
\caption{\textbf{Evaluation results on LIBERO benchmark.} We train a unified model on all suites and report the success rate on each suite.}
\label{tab:libero}
\aftertabcaption{}
\resizebox{\linewidth}{!}{
\begin{tabular}{lcccccc}
\toprule
\textbf{Method} & \textbf{Spatial} & \textbf{Object} & \textbf{Goal} & \textbf{Long} & \textbf{Average} \\
\midrule
Diffusion Policy~\cite{Chi2023DiffusionPolicy} & 78.5 & 87.5 & 73.5 & 64.8 & 76.1 \\
\rowcolor{lightgray}
OpenVLA~\cite{kim24openvla} & 84.7 & 88.4 & 79.2 & 53.7 & 76.5 \\
SpatialVLA~\cite{qu2025spatialvla} & 88.2 & 89.9 & 78.6 & 55.5 & 78.1 \\
\rowcolor{lightgray}
CoT-VLA~\cite{zhao2025cotvla} & 87.5 & 91.6 & 87.6 & 69.0 & 83.9 \\
$\pi_0$-Fast~\cite{Pertsch2025fast} & 96.4 & 96.8 & 88.6 & 60.2 & 85.5 \\
\rowcolor{lightgray}
GR00T-N1~\cite{gr00tn1} & 94.4 & 97.6 & 93.0 & 90.6 & 93.9 \\
$\pi_0$~\cite{pi0} & 98.0 & 96.8 & 94.4 & 88.4 & 94.4 \\
\rowcolor{lightgray}
F1~\cite{lv2025f1}  & 98.2 & 97.8 & 95.4 & 91.3 & 95.7 \\
InternVLA-M1~\cite{internvlam1}  & 98.0 & 99.0 & 93.8 & 92.6 & 95.9 \\
\rowcolor{lightgray}
Dis. Diff. VLA~\cite{liang2025discrete} & 97.2 & 98.6 & 97.4 & 92.0 & 96.3 \\
$\pi_{0.5}$~\cite{pi_0.5} & 98.8 & 98.2 & 98.0 & 92.4 & 96.9 \\
\rowcolor{lightgray}
GR00T-N1.6~\cite{gr00tn1} & 97.7 &98.5&97.5&94.4& 97.0\\
OpenVLA-OFT~\cite{kim2025fine} & 97.6 & 98.4 & 97.9 & 94.5 & 97.1 \\
\rowcolor{lightgray}
UniVLA~\cite{bu2025univla}&96.5& 96.8& 95.6& 92.0& 95.2\\
X-VLA~\cite{zheng2025xvla} & 98.2 & 98.6 & 97.8 & \textbf{97.6} & 98.1 \\
\rowcolor{lightgray}
GeoVLA~\cite{sun2025geovla}&98.4 &99.0 &96.6& 96.6& 97.7\\
3D-CAVLA~\cite{bhat20253d}&98.2& 99.8& 98.2 &96.1& 98.1\\
\rowcolor{lightgray}
Spatial Forcing~\cite{li2025spatial} &\textbf{99.4} &99.6 &98.8 &96.0 &98.5\\
\midrule
\textbf{Ours}  & 98.8 & \textbf{99.8} & \textbf{99.0} & 96.6 & \textbf{98.6} \\
\bottomrule
\end{tabular}
}
\end{table}

%% file: Tables/robotwin.tex
\begin{table}[t]
  \centering
  \caption{\textbf{Evaluation results on RoboTwin 2.0 benchmark.} The results are evaluated using a single model for both ``Clean'' and ``Randomized'' settings.}
  \aftertabcaption{}
  \label{tab:robotwin}
\resizebox{\linewidth}{!}{  
\begin{tabular}{ccccccc}
    \toprule
    \textbf{\multirow{2}{*}{Simulation Task}} 
      & \multicolumn{2}{c}{$\bm{\pi}_{\mathbf{0.5}}$~\cite{pi_0.5}}
      & \multicolumn{2}{c}{\textbf{X-VLA}~\cite{zheng2025xvla}}
      & \multicolumn{2}{c}{\textbf{Ours}} \\
    & \textbf{Clean} & \textbf{Rand.} 
    & \textbf{Clean} & \textbf{Rand.} 
    & \textbf{Clean} & \textbf{Rand.}  \\
    \midrule
    \textit{Place Dual Shoes} & 12  & 7  & 79  & \textbf{88}  &  \textbf{88} & 86 \\
\textit{Move Stapler Pad} & 16  & 18  & \textbf{78}  & 73  &  61 & 60 \\
\textit{Stack Blocks Two} & 48  & 56  & 92  & 87  &98 &\textbf{99} \\
\textit{Scan Object} & 42  & 38  & 14  & 36  &\textbf{91} &87  \\
\textit{Place Object Stand} & 74  & 65  & 86  & 88  & \textbf{90}&87  \\
\textit{Place Fan} & 25  & 36  & 80  & 75  &90& \textbf{94} \\
\textit{Move Pillbottle Pad} & 33  & 29  & 73  & 71  &\textbf{96} & \textbf{96} \\
\textit{Pick Dual Bottles} & 10  & 6  & 47  & 36  &\textbf{88} &76  \\
\textit{Blocks Ranking Rgb} & 43  & 35  & 83  & 83  &95 & \textbf{98} \\
\textit{......(50 tasks)} &-&-&-&-&-&- \\
\textit{Turn Switch} & 5  & 6  & 40  & 61  & 55&\textbf{72}  \\
\textit{Pick Diverse Bottles} & 5  & 3  & 58  & 36  &\textbf{80} &74  \\
\textit{Place Bread Basket} & 48  & 56  & 81  & 71  &\textbf{86} &83 \\
\textit{Stack Blocks Three} & 15  & 16  & 6  & 10  & \textbf{91}  &84 \\
\textit{Put Bottles Dustbin} & 12  & 9  & 74  & 77  & 62& \textbf{87} \\
\textit{Place Can Basket} & 19  & 25  & 49  & 52  &\textbf{82} &73 \\
\textit{Stamp Seal} & 36  & 23  & 76  & 82  &81 &\textbf{88} \\
\textit{Handover Block} & 18  & 19  & 73  & 37  & 87&\textbf{89}  \\
\textit{Stack Bowls Three} & 33  & 35  & 76  &86  &81 &\textbf{92}  \\
\textit{Place Object Basket} & 43  & 36  & 44  & 39  &85 &\textbf{93}  \\
\textit{Open Microwave} & 35  & 37  & 79  & 71   & \textbf{93}&90 \\
    \midrule
    \textbf{\textit{Average}} & 42.98 & 43.84 & 72.80 & 72.84 &85.18 &\textbf{86.06}\\
    \bottomrule
  \end{tabular}
}

\end{table}

%% file: Tables/vlm_feature.tex
\begin{table*}[htbp]\centering
\caption{\textbf{Ablation study of VLM feature interaction manners on LIBERO-Plus.} We investigate the impact of feature source layers (Last vs. Intermediate vs. Last 16) and conditioning types (Original Features vs. Learnable Queries) on action generation. }
\aftertabcaption{}
\label{tab:vlm_feature}
\resizebox{0.9\linewidth}{!}{%
\begin{tabular}{c|cc|ccccccccc}\toprule
\textbf{Layers}&\textbf{Feature}&\textbf{Query}&\textbf{Camera} &\textbf{Robot} &\textbf{Language} &\textbf{Light} &\textbf{Background} &\textbf{Noise} &\textbf{Layout} &\textbf{Total} \\
\midrule

\multirow{2}{*}{Last} & \checkmark  &$\times$&39.9&\textbf{63.7}&85.9&90.8&\textbf{93.1}&60.6&\textbf{76.8}&\textbf{71.0}  \\
 &$\times$ & \checkmark &38.4&61.5&87.2&91.6&90.3&59.1&75.3&70.0  \\
\midrule

\multirow{2}{*}{Intermediate} &  \checkmark &$\times$&38.2&49.9&88.7&92.7&89.5&\textbf{64.3}&73.6&69.0  \\
 &$\times$ & \checkmark &\textbf{42.3}&54.4&\textbf{89.1}&87.5&88.7&53.9&75.1&68.3  \\

\midrule

\multirow{3}{*}{Last 16 layers} & \checkmark  &$\times$&36.9&54.1&87.1&\textbf{93.9}&90.3&51.1&74.4&67.4  \\
 &$\times$ & \checkmark &25.8&57.5&80.9&88.7&87.5&37.3&74.1&65.2  \\
 & \checkmark  &\checkmark&25.0&50.4&88.5&90.1&88.5&50.0&70.3&63.8  \\

\bottomrule
\end{tabular}
}

\end{table*}

%% file: Tables/vlm_fusion.tex
\begin{table*}[htbp]
\centering
\caption{\textbf{Ablation study of fusion strategies between VLM and spatial feature on LIBERO-
Plus.}}
\aftertabcaption{}
\label{tab:vlm_fusion}
\resizebox{0.9\linewidth}{!}{
\begin{tabular}{lcccccccc}\toprule
\textbf{Method} & \textbf{Camera} & \textbf{Robot} & \textbf{Language} & \textbf{Light} & \textbf{Background} & \textbf{Noise} & \textbf{Layout} & \textbf{Total} \\
\midrule
Cross Attn. &\textbf{45.8}&\textbf{53.8}&86.8&\textbf{97.2}&\textbf{93.9}&\textbf{65.5}&69.8&\textbf{71.1}\\
Concat&41.2&51.2&87.0&93.7&88.4&63.9&70.6&68.9\\
Q-former&44.3&51.0&\textbf{87.2}&95.1&91.3&62.0&\textbf{71.0}&69.6\\
\bottomrule
\end{tabular}
}
\end{table*}

%% file: Tables/g3t.tex
\begin{table*}[htbp]
\centering
\caption{\textbf{Ablation study of $\text{G}^3\text{T}$ on LIBERO-
Plus.} 
We compare our proposed $\text{G}^3\text{T}$ against alternative fusion methods. 
``Cross Attn.'' uses VGGT features as query and LongCat-Image-Edit features as key/value, while ``Inv. Cross Attn.'' reverses this role. }
\aftertabcaption{}
\label{tab:g3t}

\resizebox{0.9\linewidth}{!}{
\begin{tabular}{lcccccccc}\toprule
\textbf{Method} & \textbf{Camera} & \textbf{Robot} & \textbf{Language} & \textbf{Light} & \textbf{Background} & \textbf{Noise} & \textbf{Layout} & \textbf{Total} \\
\midrule
Concat&51.9&\textbf{62.5}&85.1&89.5&91.7&77.4&81.9&75.8\\
Cross Attn.&59.5&56.8&89.0&92.6&92.3&79.8&75.1&76.5\\
Inv. Cross Attn.&56.8&56.6&\textbf{89.4}&93.0&91.7&75.1&73.4&75.1\\
Self Attn.&60.4&61.6&88.9&91.7&90.2&\textbf{80.6}&\textbf{76.7}&77.4\\

\midrule
\textbf{$\text{G}^3\text{T}$} & \textbf{62.6} & 60.9 & 89.3 & \textbf{94.3} & \textbf{92.6} & 78.2 & 76.6 & \textbf{77.9} \\
\bottomrule
\end{tabular}
}
\end{table*}

%% file: Tables/ablation.tex
\begin{table}[t]
\centering
\caption{\textbf{Ablation study of key components on LIBERO-Plus.} ``LongCat'' refers to LongCat-Image-Edit model.}
\aftertabcaption{}
\label{tab:ablation}

\resizebox{\linewidth}{!}{
\begin{tabular}{ccccc}
\toprule
 \textbf{VGGT} & \textbf{LongCat (1 view)} &\textbf{LongCat (2 views)} & \textbf{AML} &\textbf{Result} \\
\midrule
&&&&66.4\\
\rowcolor{lightgray} 
\checkmark&&&&71.1\\
&\checkmark&&&68.0\\
\rowcolor{lightgray} 
&&\checkmark&&70.2\\
&&&\checkmark&72.4\\
\rowcolor{lightgray} 
\checkmark&&\checkmark&&77.9\\
\checkmark&&\checkmark&\checkmark&\textbf{85.7}\\

\bottomrule
\end{tabular}
}
\end{table}

%% file: Tables/depth.tex
\begin{table}[t]
    \centering
    \caption{\textbf{Quantitative comparison of depth estimation on LIBERO.} We report Absolute Relative Error (AbsRel), Root Mean Squared Error (RMSE), and Log10 RMSE (lower is better $\downarrow$), as well as accuracy thresholds $\delta < 1.25^k$ for $k=1,2,3$ (higher is better $\uparrow$).}
    \aftertabcaption{}
    \label{tab:depth}
    \resizebox{\linewidth}{!}{%
    \begin{tabular}{lcccccc}
        \toprule
        \textbf{Method} & \textbf{AbsRel}$  \downarrow$ & \textbf{RMSE} $\downarrow$ & \textbf{Log10} $\downarrow$ & $\mathbf{\bm{\delta}_1}$ $\uparrow$ & $\mathbf{\bm{\delta}_2}$ $\uparrow$ & $\mathbf{\bm{\delta}_3}$ $\uparrow$ \\
        \midrule
        VGGT   & 3.5614  & 0.5557 & 0.5291 & 0.0265    & 0.1243    & 0.4810    \\
        $\text{G}^3\text{T}$ (Ours)   & \textbf{3.3353} & \textbf{0.4559} & \textbf{0.5112} & \textbf{0.0791} & \textbf{0.3373} & \textbf{0.5061} \\
        \bottomrule
    \end{tabular}%
    }
\end{table}

%% file: Tables/robust.tex

\begin{table}[t]
\centering
\caption{\textbf{Robustness comparison.} Performance degradation ($\downarrow$) under input perturbations. Our method achieves the highest original performance and the lowest degradation, demonstrating superior robustness.}
\aftertabcaption{}
\label{tab:robust_colored}

\resizebox{\linewidth}{!}{  
\begin{tabular}{lcccccc}
\toprule
\textbf{Method} & \textbf{Type} & \textbf{Spatial} & \textbf{Object} & \textbf{Goal} & \textbf{Long} & \textbf{Avg.} \\
\midrule
OpenVLA~\cite{kim24openvla} 
    & Orig. & 84.7 & 88.4 & 79.2 & 53.7 & 76.5 \\
    & Pert. & 19.4 & 14.0 & 15.1 & 14.3 & 15.6 \\
    & $\Delta$ & \color{red}$\downarrow$ 65.3 & \color{red}$\downarrow$ 74.4 & \color{red}$\downarrow$ 64.1 & \color{red}$\downarrow$ 39.4 & \color{red}$\downarrow$ 60.9 \\
\midrule
OpenVLA-OFT~\cite{kim2025fine} 
    & Orig. & 97.6 & 98.4 & 97.9 & 94.5 & 97.1 \\
    & Pert. & 84.0 & 66.5 & 63.0 & 66.4 & 69.6 \\
    & $\Delta$ & \color{orange}$\downarrow$ 13.6 & \color{red}$\downarrow$ 31.9 & \color{red}$\downarrow$ 34.9 & \color{red}$\downarrow$ 28.1 & \color{red}$\downarrow$ 27.5 \\
\midrule
$\pi_0$~\cite{pi0} 
    & Orig. & 98.0 & 96.8 & 94.4 & 88.4 & 94.4 \\
    & Pert. & 60.7 & 61.4 & 44.9 & 48.4 & 53.6 \\
    & $\Delta$ & \color{red}$\downarrow$ 37.3 & \color{red}$\downarrow$ 35.4 & \color{red}$\downarrow$ 49.5 & \color{red}$\downarrow$ 40.0 & \color{red}$\downarrow$ 40.8 \\
\midrule
$\pi_0$-Fast~\cite{Pertsch2025fast} 
    & Orig. & 96.4 & 96.8 & 88.6 & 60.2 & 85.5 \\
    & Pert. & 74.4 & 72.7 & 57.5 & 43.4 & 61.6 \\
    & $\Delta$ & \color{orange}$\downarrow$ 22.0 & \color{orange}$\downarrow$ 24.1 & \color{red}$\downarrow$ 31.1 & \color{orange}$\downarrow$ 16.8 & \color{red}$\downarrow$ 23.9 \\
\midrule
UniVLA~\cite{bu2025univla} 
    & Orig. & 96.5 & 96.8 & 95.6 & 92.0 & 95.2 \\
    & Pert. & 55.5 & 36.7 & 40.7 & 36.9 & 42.9 \\
    & $\Delta$ & \color{red}$\downarrow$ 41.0 & \color{red}$\downarrow$ 60.1 & \color{red}$\downarrow$ 54.9 & \color{red}$\downarrow$ 55.1 & \color{red}$\downarrow$ 
    52.3 \\
\midrule
\rowcolor{lightblue} \textbf{Ours} &  Orig. & \textbf{98.8} & \textbf{99.8} &\textbf{99.0} & \textbf{96.6} & \textbf{98.6} \\
\rowcolor{lightblue} 
    &  Pert. & \textbf{90.8} & \textbf{86.5} & \textbf{82.4} & \textbf{83.5} & \textbf{85.7} \\
\rowcolor{lightblue} &  
    $\Delta$ & 
    \textcolor{green!50!black}{$\downarrow$ \textbf{8.0}} & 
    \textcolor{green!50!black}{$\downarrow$ 13.3} & 
    \textcolor{green!50!black}{$\downarrow$ 16.6} & 
    \textcolor{green!50!black}{$\downarrow$ 13.1} & 
    \textcolor{green!50!black}{$\downarrow$ \textbf{12.9}} \\
\bottomrule
\end{tabular}
}

\end{table}

%% file: sec/sec4_experiments.tex
\section{Experiments}
\label{sec:exp}

\input{Tables/aml}

\input{Figures/depth}
\input{Figures/gate}
\input{Figures/real_setup}

To rigorously evaluate the efficacy of our proposed framework, we conduct extensive experiments across both simulation and real-world environments. Specifically, we aim to address the following key research questions: 
(1) How does our method compare against state-of-the-art VLA baselines? 
(2) What is the individual contribution of each proposed module to the overall performance? 
(3) Can $\text{G}^3\text{T}$ effectively resolve depth ambiguity and adaptively select informative views? 
(4) How robust is our method against external disturbances compared to baseline models? 
(5) Does the Action Manifold Learning (AML) module enhance the efficiency of action learning? 
(6) How well does our method perform in real-world robot setups? 
(7) To what extent does our method generalize to unseen tasks in real-world scenarios?

\subsection{Comparison with State-of-the-Art Methods}

To comprehensively evaluate the effectiveness of our proposed framework, we conduct extensive comparative experiments against state-of-the-art VLA baselines across three representative simulation benchmarks: LIBERO~\cite{liu2023libero}, LIBERO-Plus~\cite{fei2025libero_plus}, and RoboTwin 2.0~\cite{chen2025robotwin}. These benchmarks are selected to rigorously assess our method's performance in standard manipulation tasks, robustness against environmental perturbations, and generalization in complex bi-manual scenarios, respectively. 

Specifically, LIBERO~\cite{liu2023libero} serves as the benchmark for evaluating policy generalization across variations in spatial layouts, object instances, goal specifications, and long-horizon dependencies. It comprises four distinct suites (Spatial, Object, Goal, and Long), each containing 10 diverse tasks with 500 expert demonstrations per task. To test robustness in unseen conditions, we utilize LIBERO-Plus~\cite{fei2025libero_plus}, which extends the original LIBERO by introducing controllable perturbations across seven dimensions, including camera viewpoints, lighting, background textures, and visual noise. This allows for a zero-shot evaluation of the model's inherent resilience to sensory and environmental disturbances. Finally, we employ RoboTwin 2.0~\cite{chen2025robotwin} to assess performance in challenging bi-manual manipulation tasks. This benchmark involves training on a mix of clean and heavily randomized scenes (featuring random backgrounds, clutter, and table heights), thereby testing the model's capability to handle complex multi-arm coordination under significant domain randomization.

The quantitative results of these comprehensive evaluations are summarized in Tab.~\ref{tab:libero_plus}, Tab.~\ref{tab:libero}, and Tab.~\ref{tab:robotwin}. Our method consistently outperforms existing state-of-the-art baselines across all three benchmarks. Specifically, it achieves an average success rate of 98.6\% on LIBERO, demonstrates superior zero-shot robustness with an 85.7\% success rate on LIBERO-Plus, and attains over 80\% success in the complex bi-manual settings of RoboTwin 2.0. These results collectively validate the superiority of our method in terms of task success rate, robustness, and generalization capability.

\subsection{Ablation Studies}
\label{subsec:ablation}

\vspace{0.5em}
\noindent \textbf{Analysis of VLM feature interaction.}
To determine the optimal VLM feature extraction and conditioning strategy, we compare different feature sources (Last layer, Intermediate layers, or Average of Last 16 layers) and conditioning mechanisms (Direct usage vs. Learnable Action Queries). This experiment is conducted excluding spatial features (VGGT and multi-view priors) and the AML module. As summarized in Tab.~\ref{tab:vlm_feature}, using final-layer features with direct conditioning yields the best performance. Other combinations, such as intermediate features or hybrid conditioning, result in lower success rates due to noisy low-level details or representation conflicts.

\vspace{0.5em}
\noindent \textbf{Impact of feature fusion strategy.}
We evaluate different methods for fusing semantic VLM embeddings with spatial features. The compared baselines include simple concatenation, Q-Former, and direct cross-attention. This study is conducted without the AML module to isolate the effect of the fusion mechanism. As shown in Tab.~\ref{tab:vlm_fusion}, direct cross-attention outperforms both concatenation and Q-Former, demonstrating its effectiveness in aligning cross-modal representations. Consequently, we adopt cross-attention as our standard fusion module.

\vspace{0.5em}
\noindent \textbf{Effectiveness of $\text{G}^3\text{T}$ fusion strategy.}
To justify the design of our $\text{G}^3\text{T}$ module, we compare it against static fusion alternatives, including simple concatenation, standard cross-attention (VGGT as query), and inverse cross-attention. Results in Tab.~\ref{tab:g3t} show that static methods struggle to handle occlusion and view ambiguity. In contrast, our proposed $\text{G}^3\text{T}$, which employs adaptive gating based on geometric consistency, achieves the highest success rate, validating the necessity of dynamic feature aggregation.

\vspace{0.5em}
\noindent \textbf{Impact of key components.}
Finally, we perform a step-by-step ablation on the LIBERO-Plus benchmark to assess the contribution of each key component in Tab.~\ref{tab:ablation}. The baseline consists of monocular visual features and the standard GR00T-N1 action head. We sequentially add: (1) VGGT feature for 3D geometric prior, (2) Multi-view features from LongCat-Image-Edit, and (3) our proposed Action Manifold Learning (AML) module. Each addition leads to a consistent performance gain. The full framework, integrating $\text{G}^3\text{T}$ and AML, achieves the highest success rate, significantly outperforming the baseline and confirming the effectiveness of all proposed modules.

\input{Tables/real}
\input{Figures/real_demo}
\input{Figures/zeroshot_setup}
\input{Figures/zeroshot_demo}

\subsection{Analysis of $\text{G}^3\text{T}$}
\label{subsec:g3t_analysis}

Following the ablation studies, we further investigate the internal mechanisms of $\text{G}^3\text{T}$ to address two key questions: (1) Does view synthesis effectively resolve depth ambiguity in monocular input? and (2) Does the gating mechanism adaptively suppress noise while selecting informative geometric cues?

\vspace{0.5em}
\noindent \textbf{Depth estimation analysis.}
To verify if $\text{G}^3\text{T}$ enhances spatial perception, we conduct a controlled depth estimation experiment. We attach a lightweight DPT head~\cite{ranftl2021vision} to frozen backbone and compare features from standard monocular VGGT against those fused with $\text{G}^3\text{T}$'s synthesized multi-view cues. Both heads are trained under identical setting to ensure fair comparison. As shown in Tab.~\ref{tab:depth}, $\text{G}^3\text{T}$ consistently outperforms the monocular baseline across all metrics. The results confirm that synthesized multi-view cues provide more consistent geometric priors, effectively resolving depth ambiguities inherent in single-view input. These quantitative gains align with the qualitative results in Fig.~\ref{fig:depth}, where $\text{G}^3\text{T}$ yields sharper object boundaries and more coherent depth structures.

\vspace{0.5em}
\noindent \textbf{Visualization of adaptive gating.}
While synthesized views enrich geometric context, they may introduce artifacts. The adaptive gating mechanism in $\text{G}^3\text{T}$ is designed to distinguish reliable signals from such noise. Fig.~\ref{fig:gate} visualizes the learned gate maps, where warmer colors indicate high confidence and cooler colors denote uncertainty. The gating mechanism demonstrates strong semantic and geometric awareness. High gate values concentrate on reliable structures, such as object boundaries, distinct textures, and visible surfaces. Conversely, low values correspond to regions with severe occlusion, motion blur, or synthesis artifacts. This validates that $\text{G}^3\text{T}$ does not blindly fuse information but constructs a robust geometric representation by attending only to the most trustworthy cues.

\subsection{Robustness to Environmental Perturbations}
\label{subsec:robustness}

We evaluate robustness by comparing performance on standard LIBERO (Orig.) against the perturbed LIBERO-Plus (Pert.), which introduces diverse environmental variations. As shown in Tab.~\ref{tab:robust_colored}, our method achieves the highest original accuracy with the lowest average performance degradation, significantly outperforming baselines like OpenVLA-OFT  and $\pi_0$. Notably, we observe minimal degradation in Spatial tasks (8.0\%), validating that our geometric-aware architecture effectively anchors policy decisions to stable 3D structures, thereby ensuring reliability under severe visual perturbations.

\subsection{Efficiency of Action Manifold Learning}
\label{subsec:aml_efficiency}

We compare AML against the GR00T baseline under identical settings (Qwen3-VL backbone, 0.16B action expert, no auxiliary 3D modules), varying denoising steps and action chunk sizes to assess efficiency and scalability. As shown in Tab.~\ref{tab:aml}, AML consistently outperforms GR00T, maintaining robust performance even with minimal sampling steps, which indicates a more structured optimization landscape. Furthermore, while GR00T suffers from severe degradation as action chunk size increases, AML exhibits graceful degradation. This demonstrates that AML effectively captures stable geometric correlations, ensuring superior efficiency and robustness in high-dimensional temporal predictions without excessive computational cost.

\input{Tables/zeroshot_clean}
\input{Tables/zeroshot_cluttered}

\subsection{Real-world Experiment}
\label{subsec:real}

\vspace{0.5em}
\noindent \textbf{Setup.}
We design four distinct manipulation tasks to evaluate the robot's performance across varying geometric constraints and stability requirements using a single Franka Emika Panda arm. The tasks include \textit{stack block} (Stack the blue block on the red block), \textit{insert cube} (Insert the pink cube into the red cup), \textit{place cylinder} (Place the blue cylinder on the green block) and \textit{place cup} (Place the yellow cup on the red block). Figure~\ref{fig:real_setup} illustrates the experimental setup for these four tasks. For data collection, we record 20 demonstration episodes per task. Each model is evaluated over 10 independent trials per task, with the success rate serving as the primary metric. We jointly fine-tune a single unified model on all tasks for $30K$ steps.

\vspace{0.5em}
\noindent \textbf{Results.}
We compare our unified model against the representative VLA baselines, OpenVLA-OFT~\cite{kim2025fine} and $\pi_0$~\cite{pi0}. To ensure a fair comparison, we fine-tune them on the same mixed dataset using identical training protocols. As shown in Table~\ref{tab:real}, our method outperforms the baselines across all four tasks. These results demonstrate that our method enables effective deployment in diverse real-world manipulation scenarios. The qualitative results are shown in Fig~\ref{fig:real_demo}. Additional video demonstrations of our experimental results are provided in our project page.

\subsection{Zero-shot generalization}
\label{subsec:generalization}
To evaluate the zero-shot generalization capability of our model, we test it on unseen tasks that require compositional reasoning beyond the training distribution. Specifically, we modify the object attributes in the language instructions while keeping the manipulation primitive unchanged. The four evaluation tasks are: \textit{``Insert the pink cube into the blue cup''}, \textit{``Place the blue cylinder on the red block''}, \textit{``Place the orange cup on the red block''}, and \textit{``Stack the red block on the blue/green block''}. Note that these specific color-object combinations were not present in the training set.

We categorize the evaluation scenarios into two difficulty levels based on visual complexity:
(1) Clean Context: The workspace contains only the target objects specified in the instruction. This setting isolates the model's ability to generalize semantic concepts.
(2) Cluttered Context: The workspace includes the target objects along with distractor objects seen during training (e.g., other blocks or cups). This setting challenges the model's robustness against visual interference and its ability to attend to the correct objects amidst clutter.

Fig.~\ref{fig:zeroshot_setup} illustrates the three typical scenarios: a training example, a clean context test case, and a cluttered context test case. We evaluate the model over 10 independent trials for each unseen task variant. Tab.~\ref{tab:zeroshot_clean} and~\ref{tab:zeroshot_cluttered} report the success rates for the Clean and Cluttered contexts, respectively. We also show qualitative results of our method in Fig.~\ref{fig:zeroshot_demo}. More video results can be found in our project page.

\subsection{Limitations and Future Work}
\label{subsec:limitations}
While our method enhances 3D spatial perception and efficient action optimization, it still faces challenge regarding computational efficiency. The integration of a multi-view diffusion model for view synthesis introduces substantial overhead. Although we minimize sampling steps and resolution, and employ pre-computed multi-view features to accelerate training, the iterative generation process during inference still prevents real-time responsiveness. This latency bottleneck suggests that online synthesis is currently impractical for high-frequency control tasks. In the future, we will focus on distilling the geometric reasoning capabilities of the diffusion model directly into the VLA backbone, thereby endowing the policy with inherent 3D awareness without relying on external generative modules during deployment.

%% file: Tables/aml.tex
\begin{table*}[t]\centering

\caption{\textbf{Efficiency and robustness analysis of Action Manifold Learning (AML).} 
We evaluate the efficiency of AML compared to GR00T's velocity prediction paradigm by varying denoising steps and action chunk sizes. }
\aftertabcaption{}
\resizebox{0.9\linewidth}{!}{

\begin{tabular}{lccccccccccc}
\toprule
\multirow{2}{*}{\textbf{Method}}&\textbf{Denoising}&\textbf{Action}&\multirow{2}{*}{\textbf{Camera}} &\multirow{2}{*}{\textbf{Robot}} &\multirow{2}{*}{\textbf{Language}} &\multirow{2}{*}{\textbf{Light}} &\multirow{2}{*}{\textbf{Background}} &\multirow{2}{*}{\textbf{Noise}} &\multirow{2}{*}{\textbf{Layout}} &\multirow{2}{*}{\textbf{Total}} \\
&\textbf{Steps}&\textbf{Chunk}&&&&&&&&\\
\midrule

GR00T& \multirow{2}{*}{4} & \multirow{2}{*}{8}&41.0&61.2&86.5&91.3&91.6&53.8&73.4&69.3 \\
Ours&&&39.9&63.7&85.9&90.8&93.1&60.6&76.8&71.0  \\
\midrule
GR00T&&&35.4&59.7&86.7&90.2&90.1&49.8&73.3&67.2 \\
Ours&\multirow{-2}{*}{2} &\multirow{-2}{*}{8}&37.4&59.6&82.7&\textbf{93.1}&92.9&56.7&\textbf{80.5}&69.7  \\
\rowcolor{lightblue}
GR00T&&&36.7&61.0&88.0&91.9&91.9&51.1&74.6&68.6 \\
\rowcolor{lightblue}
Ours&\multirow{-2}{*}{10} &\multirow{-2}{*}{8} &\textbf{44.6}&56.9&81.4&90.2&90.1&62.9&78.3&70.2  \\
\midrule
GR00T&\multirow{2}{*}{4}&\multirow{2}{*}{10} &37.3&61.6&\textbf{88.7}&92.8&92.8&51.7&75.2&69.3 \\
Ours&&&42.5&\textbf{64.1}&86.3&92.7&\textbf{93.3}&\textbf{63.2}&78.0&\textbf{72.4}  \\
\rowcolor{lightblue}
GR00T&&&7.9&31.7&64.9&83.2&73.6&24.0&55.2&45.7 \\
\rowcolor{lightblue}
$\Delta$ (Change) &  &               & \textcolor{gray}{-33.1} & \textcolor{gray}{-29.5} & \textcolor{gray}{-21.6} & \textcolor{gray}{-8.1} & \textcolor{gray}{-18.0} & \textcolor{gray}{-29.8} & \textcolor{gray}{-18.2} & \textcolor{red}{-23.6} \\
\rowcolor{lightblue}
Ours&&&23.5&53.9&74.6&92.8&91.4&53.1&68.7&62.8   \\
\rowcolor{lightblue}
$\Delta$ (Change) &\multirow{-4}{*}{4}  &  \multirow{-4}{*}{30}             & \textcolor{gray}{-16.4} & \textcolor{gray}{-9.8} & \textcolor{gray}{-11.3} & \textcolor{gray}{+2.0} & \textcolor{gray}{-1.7} & \textcolor{gray}{-7.5} & \textcolor{gray}{-8.1} & \textcolor{OliveGreen}{-8.2} \\
\bottomrule
\end{tabular}
}

\label{tab:aml}
\end{table*}

%% file: Figures/depth.tex
\begin{figure}[t]
\begin{center}
\includegraphics[width=1\linewidth]{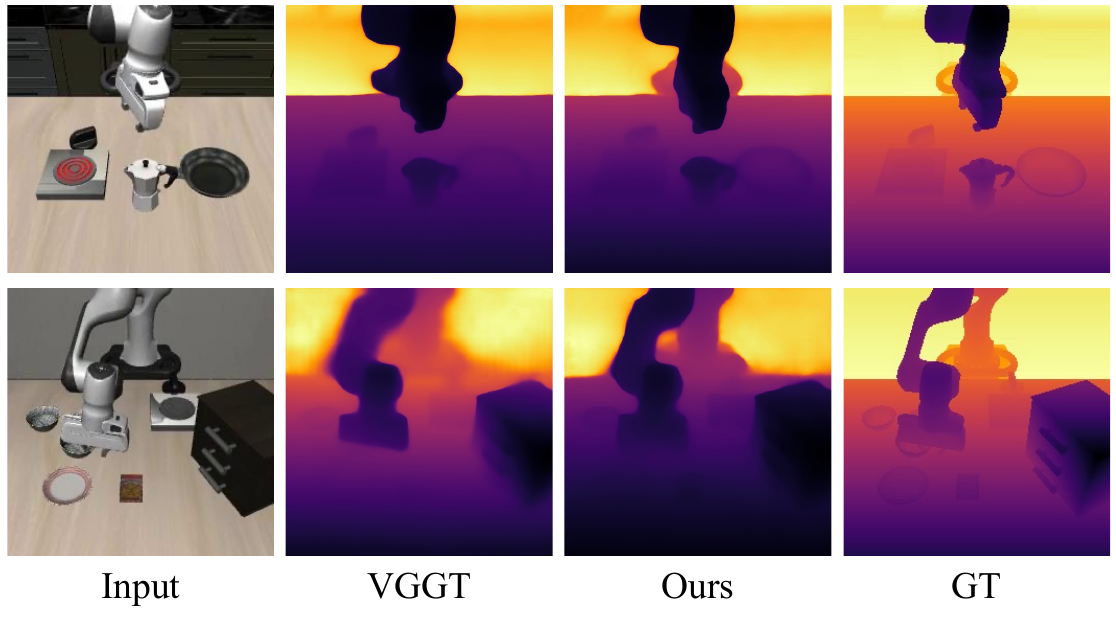} \\
\end{center}
\afterfig{}
\caption{\textbf{Qualitative depth visualization.} Benefiting from our multi-view latent representations and $\text{G}^3\text{T}$ module, our spatial feature yields more robust depth estimation characterized by sharp edges and consistent spatial geometry, outperforming standard monocular baselines.}
\label{fig:depth}
\end{figure}

%% file: Figures/gate.tex
\begin{figure}[t]
\begin{center}
\includegraphics[width=1\linewidth]{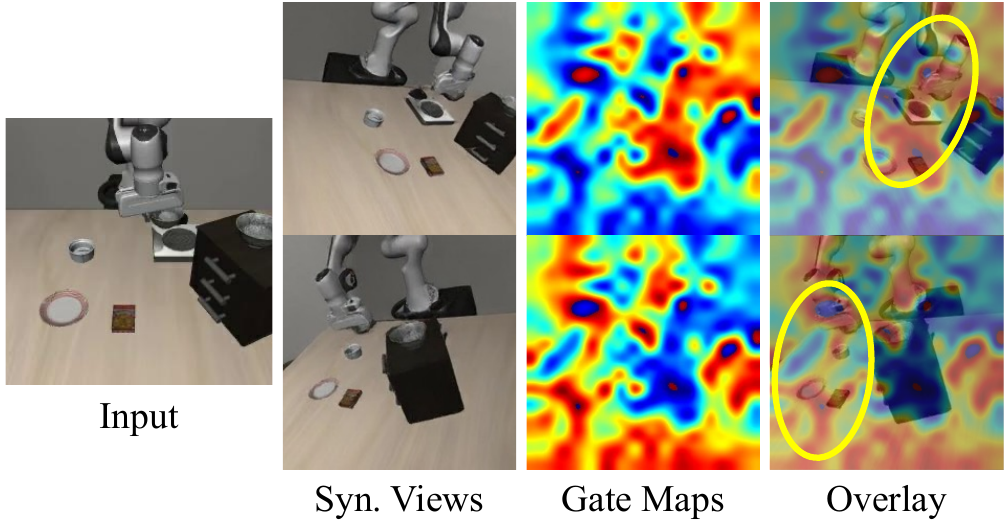} \\
\end{center}
\afterfig{}
\caption{\textbf{Visualization of $\text{G}^3\text{T}$ gating mechanism.} The gating mechanism effectively highlights reliable geometric structures (e.g., object boundaries) while suppressing uncertain regions. }
\label{fig:gate}
\end{figure}

%% file: Figures/real_setup.tex
\begin{figure}[t]
\begin{center}
\includegraphics[width=1.0\linewidth]{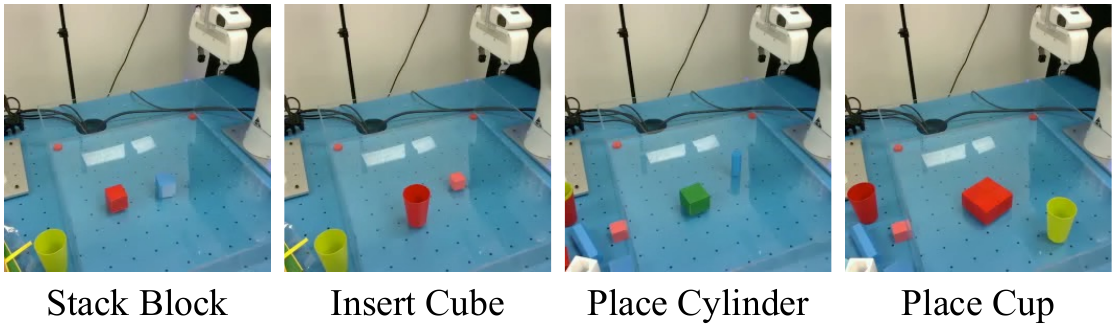} \\
\end{center}
\afterfig{}
\caption{\textbf{Real-world experimental setup using the Franka Emika Panda robot.} }
\label{fig:real_setup}
\end{figure}

%% file: Tables/real.tex
\begin{table}[t]
\centering
\caption{\textbf{Real-world comparison.} We train a single unified model on all four tasks and evaluate on each task over 10 independent trials.}
\aftertabcaption{}
\label{tab:real}
\resizebox{\linewidth}{!}{
\begin{tabular}{lcccc}
\toprule
\textbf{Method} & \textbf{Stack Block} & \textbf{Insert Cube} & \textbf{Place Cylinder} & \textbf{Place Cup} \\
\midrule
OpenVLA-OFT~\cite{kim2025fine} & 60 &  40&  30& 30 \\
$\pi_0$~\cite{pi0} & 50 &  40&  20& 20 \\
Ours  & 70 & 60 & 60 & 70  \\
\bottomrule
\end{tabular}
}
\end{table}

%% file: Figures/real_demo.tex
\begin{figure*}[t]
\begin{center}
\includegraphics[width=1\linewidth]{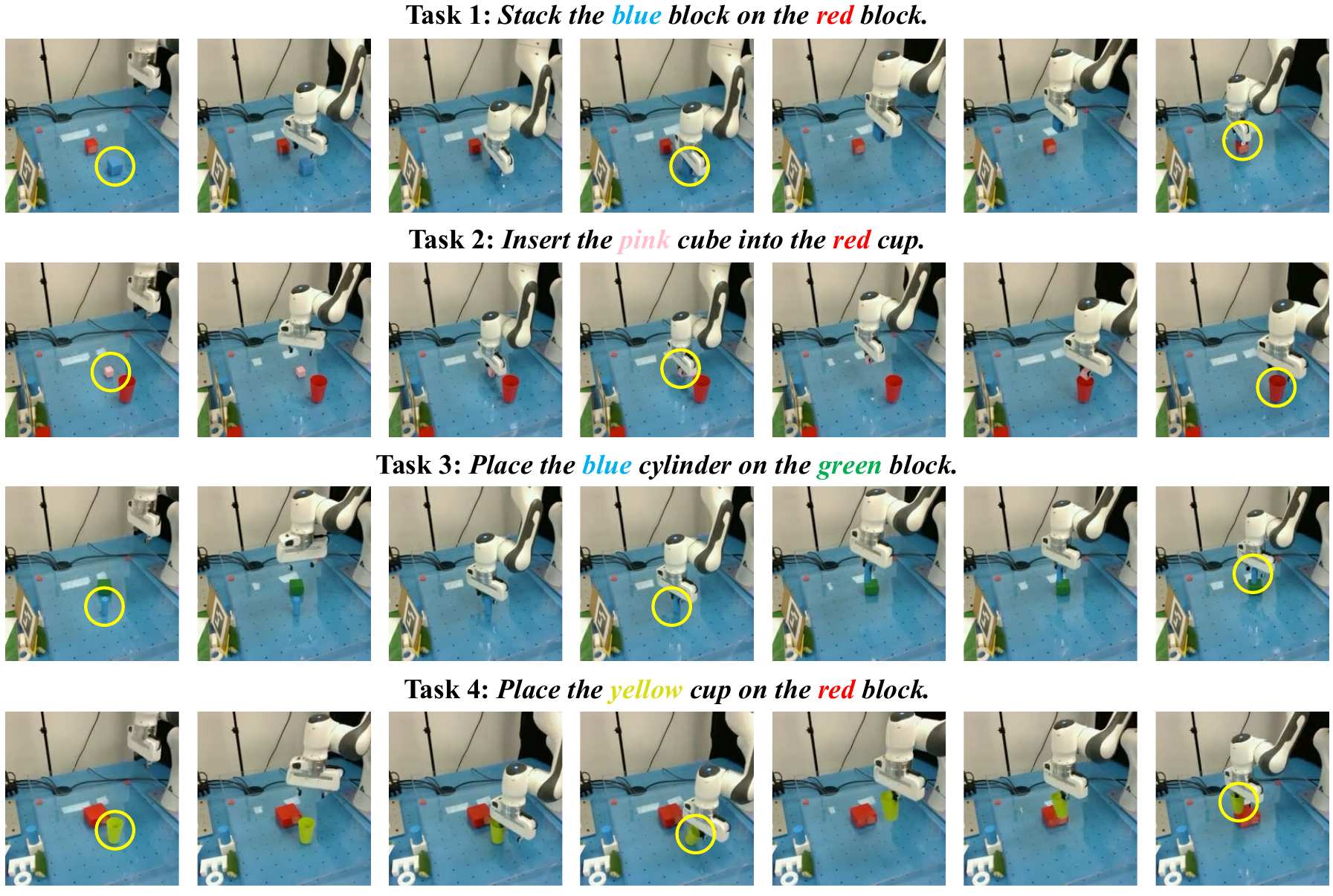} \\
\end{center}
\afterfig{}
\caption{\textbf{Qualitative results of our method in trained context.}}
\label{fig:real_demo}
\end{figure*}

%% file: Figures/zeroshot_setup.tex
\begin{figure}[t]
    \centering
    \includegraphics[width=1\linewidth]{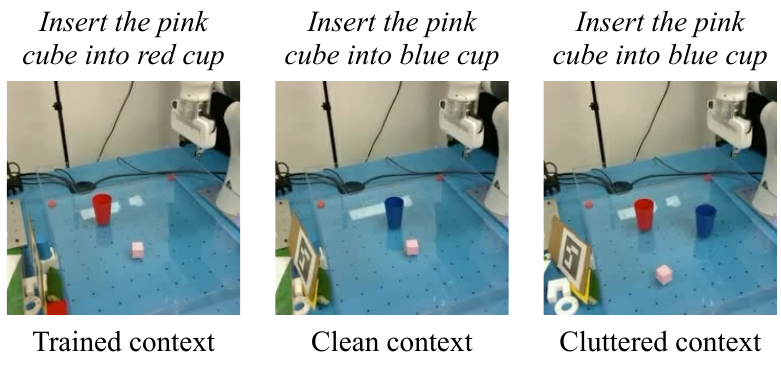} 
    \afterfig{}
    \caption{\textbf{Zero-shot generalization setup.} 
    }
    \label{fig:zeroshot_setup}
\end{figure}

%% file: Figures/zeroshot_demo.tex
\begin{figure*}[htbp]
\begin{center}
\includegraphics[width=1.0\linewidth]{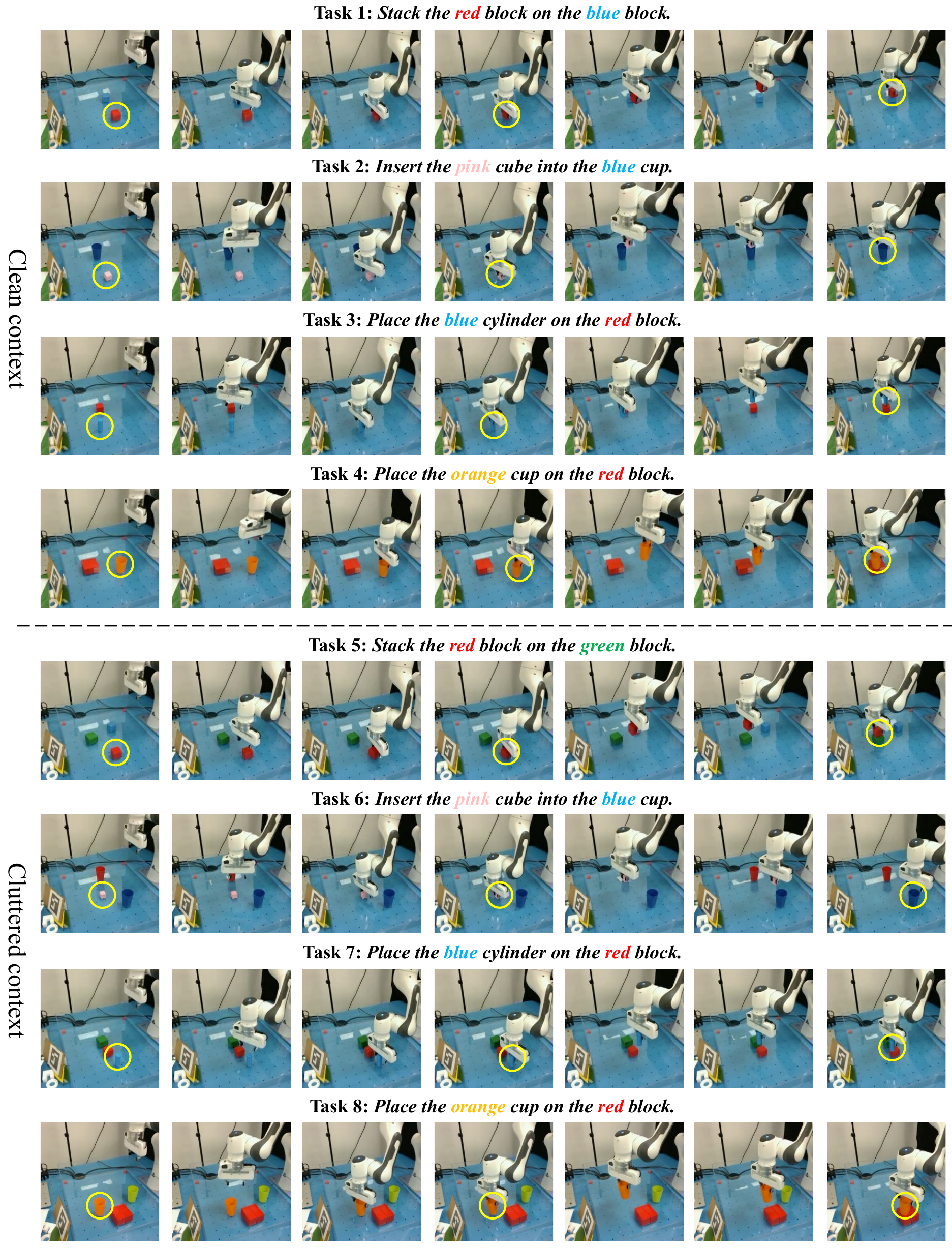} 
\end{center}
\afterfig{}
\caption{\textbf{Qualitative results of our method in clean and cluttered context.}}
\label{fig:zeroshot_demo}
\end{figure*}

%% file: Tables/zeroshot_clean.tex
\begin{table}[t]
\centering
\caption{\textbf{Zero-shot generalization in clean context.} Success rates on unseen object attributes with no visual distractors.}
\aftertabcaption{}
\label{tab:zeroshot_clean}
\resizebox{\linewidth}{!}{
\begin{tabular}{lcccc}
\toprule
\textbf{Method} & \textbf{Stack Block} & \textbf{Insert Cube} & \textbf{Place Cylinder} & \textbf{Place Cup} \\
\midrule
OpenVLA-OFT~\cite{kim2025fine} & 30 & 50 & 30 &30  \\
$\pi_0$~\cite{pi0} & 30 &  40&  20& 20 \\
Ours  & 50&60 & 50 &70  \\
\bottomrule
\end{tabular}
}
\end{table}

%% file: Tables/zeroshot_cluttered.tex
\begin{table}[t]
\centering
\caption{\textbf{Zero-shot generalization in cluttered context.} Success rates on unseen object attributes with visual distractors present.}
\aftertabcaption{}
\label{tab:zeroshot_cluttered}
\resizebox{\linewidth}{!}{
\begin{tabular}{lcccc}
\toprule
\textbf{Method} & \textbf{Stack Block} & \textbf{Insert Cube} & \textbf{Place Cylinder} & \textbf{Place Cup} \\
\midrule
OpenVLA-OFT~\cite{kim2025fine} & 30 & 10 & 20 & 10 \\
$\pi_0$~\cite{pi0} & 10 &  20&  10& 10 \\
Ours  & 50 & 40 & 30  &40  \\
\bottomrule
\end{tabular}
}
\end{table}

%% file: sec/sec5_conclusion.tex
\section{Conclusion}
\label{sec:conclusion}
We have presented a VLA framework that addresses the challenges of monocular depth ambiguity and inefficient action learning by integrating geometrically consistent view synthesis with direct action decoding. To this end, we develop Geometry-Guided Gated Transformer ($\text{G}^3\text{T}$) that leverages multi-view latent priors and an adaptive gating mechanism to resolve spatial ambiguities and filter occlusion noise, enhancing 3D perception without utilizing additional hardware. Besides, we introduce Action Manifold Learning (AML), a paradigm shifts from indirect noise/velocity prediction to direct action decoding on a low-dimensional manifold, allowing better optimization efficiency. Extensive evaluations on LIBERO, LIBERO-Plus, RoboTwin 2.0, and real-world robot experiments demonstrate that our method consistently outperforms state-of-the-art baselines in both success rate and robustness.
